%% file: main.tex
\documentclass[10pt,twocolumn,letterpaper]{article}

\usepackage{cvpr}              
\usepackage{times}
\usepackage{epsfig}
\usepackage{graphicx}
\usepackage{amsmath}
\usepackage{amssymb}
\usepackage{pifont}
\usepackage{booktabs}   
\usepackage{arydshln}   
\usepackage{adjustbox}  
\usepackage{multirow}   
\usepackage{caption}
\usepackage{float}
\usepackage{color}
\usepackage{subcaption}
\usepackage[accsupp]{axessibility} 

\newcommand\vcenterhead[1]
  {%
    \multirow{2}{*}[-0.5\dimexpr\aboverulesep+\belowrulesep+\cmidrulewidth\relax]{#1}%
  }
\input{preamble}

\definecolor{cvprblue}{rgb}{0.21,0.49,0.74}
\usepackage[pagebackref,breaklinks,colorlinks,citecolor=cvprblue]{hyperref}

\title{GAFusion: Adaptive Fusing LiDAR and Camera with Multiple Guidance for \\
3D Object Detection}
\author{  Xiaotian Li$^{1*}$ \quad Baojie Fan$^{1}$\thanks{Equal Contribution.}  \thanks{Corresponding Author.} \quad Jiandong Tian $^{2}$ \quad Huijie Fan$^{2}$  \\
$^1$Nanjing University of Posts and Telecommunications \\ \quad$^2$Shenyang Institute of Automation
Chinese Academy of Science\\
{\tt\small \{xiaotianli981, jobfbj\}@gmail.com}\quad{\tt\small \{tianjd, fanhuijie\}@sia.cn}}\vspace{-0.4cm}

\begin{document}
\maketitle

\input{sec/0_abstract}    
\input{sec/1_intro}
\input{sec/2_Related_Work}

\input{sec/3_Method}

\input{sec/4_Experiments}

\input{sec/5-Conclusion}

{
    \small
    \bibliographystyle{ieeenat_fullname}
    \bibliography{main}
}

\end{document}

%% file: preamble.tex
\usepackage[dvipsnames]{xcolor}
\usepackage{booktabs}


%% file: sec/0_abstract.tex
\begin{abstract}
Recent years have witnessed the remarkable progress of 3D multi-modality object detection methods based on the Bird's-Eye-View (BEV) perspective. However, most of them overlook the complementary interaction and guidance between LiDAR and camera. In this work, we propose a novel multi-modality 3D objection detection method, named GAFusion, with LiDAR-guided global interaction and adaptive fusion. Specifically, we introduce sparse depth guidance (SDG) and LiDAR occupancy guidance (LOG) to generate 3D features with sufficient depth information. In the following, LiDAR-guided adaptive fusion transformer (LGAFT) is developed to adaptively enhance the interaction of different modal BEV features from a global perspective. Meanwhile, additional downsampling with sparse height compression and multi-scale dual-path transformer (MSDPT) are designed to enlarge the receptive fields of different modal features. Finally, a temporal fusion module is introduced to aggregate features from previous frames. GAFusion achieves state-of-the-art 3D object detection results with 73.6$\%$ mAP and 74.9$\%$ NDS on the nuScenes test set. 
\end{abstract}

%% file: sec/1_intro.tex
\section{Introduction}
\label{sec:intro}

3D object detection is a crucial task in autonomous driving. To cope with the complex road scenarios, multiple sensors (LiDARs or cameras) are usually employed for scene understanding. LiDAR can generate accurate but sparse 3D point clouds, which contains precise spatial information. Images have rich semantic and texture information, but lack depth information. Therefore, a natural operation is to extensively fuse LiDAR and camera to leverage the complementarity of multi-modality information, which can enable the autonomous driving system to achieve higher accuracy and robustness.

\begin{figure}[t]
  \centering
    \includegraphics[width=\linewidth]{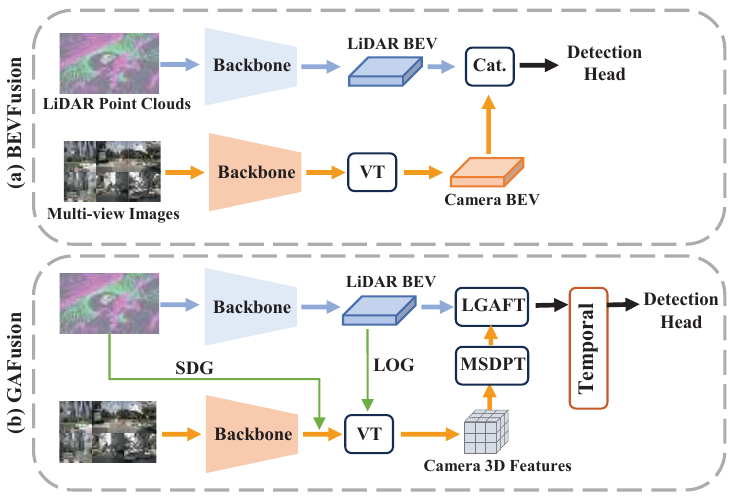}
   \caption{Comparison between BEVFusion and the proposed GAFusion. (a) In BEVFusion, the camera stream and the LiDAR stream separately generate BEV features, which are then concatenated together. (b) In GAFusion, the camera modality BEV features are generated by multiple guidance from the LiDAR stream, and the receptive fields are enhanced by MSDPT. The BEV features are fused by LGAFT. “VT” is view transformer.}\vspace{-0.2cm}
   \label{fig:comparision}
\end{figure}

Recently, fusing LiDAR and camera has achieved some progress. Early methods~\cite{chen2017mv3d, ku2018avod, Yoo20203DCVFGJ, sindagi2019mvx} achieve LiDAR-camera fusion by projecting 3D LiDAR point clouds (or region proposals) onto 2D images. But these methods overlook the information gap between the two modalities. Recent works~\cite{liangbevfusion, liu2022bevfusion, bai2022transfusion, yan2023cmt, xie2023sparsefusion, yang2022deepinteraction, jiao2022msmdfusion, cai2023bevfusion4d} adopt different query generation strategies or create a unified Bird's-Eye-View (BEV)~\cite{ma2022visionbev} intermediate feature to fuse multi-modality features. For instance, TransFusion~\cite{bai2022transfusion} applies a two-stage pipeline to fuse the camera and LiDAR features, but its performance relies on the query initialization strategy. BEVFusion~\cite{liangbevfusion, liu2022bevfusion} explores a unified representation for BEV features through view transformation, which not only preserves the spatial information of sparse LiDAR point clouds, but also lifts the 2D images to the 3D features, effectively maintaining the consistency between the two modalities. However, the camera modality still struggles with geometric perception information, which limits the complementarity between LiDAR and camera. As shown in Fig.~\ref{fig:comparision}(a), there is no interaction between both modalities.

 To tackle the above challenges, we propose an effective 3D multi-modality object detection method, named GAFusion. Within it, a LiDAR guidance module is developed, which consists of sparse depth guidance (SDG) and LiDAR occupancy guidance (LOG). SDG combines the sparse depth generated by LiDAR point clouds and the camera features to produce depth-aware features, enhancing the sensitivity of camera features to depth information. Inspired by the occupancy task~\cite{huang2023tri}, LOG guides a 3D feature volume generated by view transformation with occupancy features, and focuses on the targets in the 3D feature volume, thus providing more valuable information for fusion. Then, we construct a multi-scale dual-path transformer (MSDPT) module to improve the interactions around the objects and expand the receptive fields of the 3D feature volume. With the above designs, the camera modality has sufficient semantic features and more accurate depth distribution. In the following, to obtain abundant features in the LiDAR modality, we perform additional downsampling on the LiDAR point clouds and use sparse depth compression to aggregate features from different scales. This operation can provide larger receptive fields with less computation and memory consumption. Moreover, a LiDAR-guided adaptive fusion transformer (LGAFT) module is proposed to effectively fuse BEV features generated by LiDAR point clouds and images. In this module, the LiDAR BEV features adaptively guide the camera BEV features to strengthen the cross-modality interaction from global scope.  

All of the above operations are evaluated on single-frame raw data. In order to further explore the target correlation and motion consistency among multiple successive frames, a temporal fusion module is designed. To be specific, we store the BEV features of different frames in memory buffer, which is used to fuse the features of the previous frame and the current frame.

Our contributions are summarized as follows:

1) We propose GAFusion, a novel 3D object detection method that leverages LiDAR guidance to compensate for depth distribution of the camera features, and provides sufficient spatial information for the camera features.

2) We design LiDAR-guided adaptive fusion transformer (LGAFT), which aims to enhance the global features interaction between the two modalities in an adaptive way, facilitating the fusion of semantic and geometric features.

3) We conduct extensive experiments on the nuScenes dataset to verify the effectiveness of our GAFusion. The experiments show that without using any augmentation strategies, our model achieves the state-of-the-art performances of 72.1$\%$ mAP and 73.5$\%$ NDS.

%% file: sec/2_Related_Work.tex
\section{Related Work}
\label{sec:RW}

\begin{figure*}[htbp]
\centering
\includegraphics[width=\textwidth]{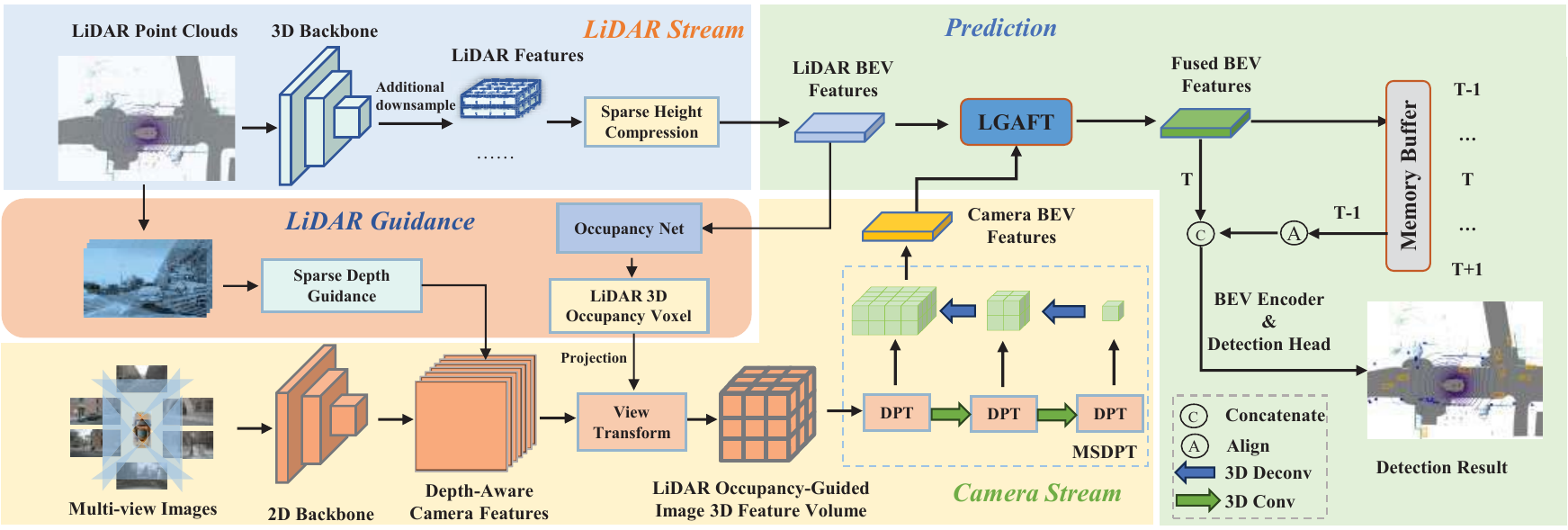}
\caption{The overall architecture of GAFusion. The multi-view images and point clouds are fed into the corresponding backbone networks to obtain multi-scale LiDAR features and camera features. For LiDAR guidance, we propose sparse depth guidance (SDG) and LiDAR occupancy guidance (LOG) to guide the 2D camera features by adopting the raw point clouds and LiDAR BEV features, respectively. In addition, we use multi-scale dual-path transformer (MSDPT) to enlarge the receptive fields. Then, LiDAR-guided adaptive fusion transformer (LGAFT) further fuses the two modalities’ BEV features. A temporal fusion module is introduced to aggregate the previous frame’s BEV features, and finally feeds these BEV features into an encoder and a detection head.}
\label{Fig.main} 
\end{figure*}
\subsection{Single-modality 3D Object Detection}
Single-modality 3D object detection, mainly including LiDAR-based 3D object detection and camera-based 3D object detection, has achieved remarkable progress in recent years. 

LiDAR-based 3D object detection aims to predict 3D object bounding boxes using the point clouds captured from LiDAR. Existing methods~\cite{qi2017pointnet, zhou2018voxelnet,lang2019pointpillars, pvrcnn, pvrcnn++, li2021lidarrcnn} either directly predict on point clouds, or convert point clouds into voxels or pillars. PointNet~\cite{qi2017pointnet} is the first framework that processes point clouds in an end-to-end manner, by taking unordered point cloud sets as direct inputs and preserving the spatial structure of point clouds. VoxelNet~\cite{zhou2018voxelnet} discretizes point clouds into voxels, and uses dense convolution to obtain BEV features.

Camera-based 3D object detection, which can be divided into two categories: image-view-based and BEV-based. DETR3D~\cite{wang2022detr3d} and PETR~\cite{liu2022petr} introduce transformer into the framework, wherein the former aggregates 2D features into 3D Query, and the latter embeds coordinate information into 2D features. They both use transformer to implicitly transform the image features to 3D space. BEVDet~\cite{huang2021bevdet} and BEVDepth~\cite{li2022bevdepth} predict the depth distribution to lift the image features to a 3D frustum meshgrid. The semantic or spatial information provided by a single-modality is still limited, despite the impressive performance achieved by the aforementioned detection tasks.
\subsection{Multi-modality 3D Object Detection}
Multi-sensor fusion has gained great attention in 3D detection due to its superior performance. Previous works~\cite{chen2017mv3d, ku2018avod, Yoo20203DCVFGJ, sindagi2019mvx} fuse 3D point cloud features and 2D image features by projecting the former onto the latter. MV3D~\cite{chen2017mv3d} associates 3D proposals with 2D RoI features and converts 3D representation into 2D pseudo images, enabling the network to leverage 2D convolutions for geometric refinement. PointPainting~\cite{pointpainting} enriches point clouds with semantic labels from images. However, the above methods fail to fully exploit the dense semantic information in images. Recently, BEVFusion~\cite{liangbevfusion, liu2022bevfusion} are proposed to fuse LiDAR features and camera features in BEV space and apply a lift-splat-shoot (LSS)~\cite{philion2020lift} operation to project image features, resulting in semantic-rich features. Before fusing BEV features from two modalities, LiDAR point clouds and images do not interact at all. From another perspective, CMT~\cite{yan2023cmt} proposes a novel end-to-end transformer-based 3D object detection framework, which implicitly encodes 3D point clouds into multi-modality tokens. Inspired by the above works, we propose the global interaction and adaptive BEV fusion that achieves significant performance improvement while maintaining the simplicity of the framework.
\subsection{Occupancy Task}
Recently, 3D occupancy prediction (Occ)~\cite{huang2023tri} has been proposed as a novel 3D detection task. Based on FB-BEV~\cite{li2023fbbev}, FB-OCC~\cite{li2023fbocc} emphasizes the importance of model scale and pre-training. OccDepth~\cite{miao2023occdepth} leverages the implicit depth information from depth images (or RGBD images) to help recover the 3D geometry. OccFormer~\cite{zhang2023occformer} utilizes a dual-path transformer for efficient long-range encoding of 3D voxel features generated by cameras. VoxFromer~\cite{li2023voxformer} adopts a two-stage design and generates a set of sparse visible and occupied voxel queries from depth estimation. OpenOccupancy~\cite{wang2023openoccupancy} is the first omnidirectional semantic occupancy perception benchmark. We notice that 3D occupancy prediction can provide more fine-grained and comprehensive 3D perception capabilities.
\subsection{Temporal Fusion}
Temporal fusion adopts multiple frames of images or point clouds to improve the performance of 3D object detection, as it can enhance the perception system’s understanding and prediction of dynamic scenes. 3D-VID~\cite{zhai20223Dvid} employs a bidirectional recurrent neural network (Bi-RNN)~\cite{schuster1997birnn} to model the temporal sequences of multiple point clouds, capturing the motion information and state changes of the targets. BEVDet4D~\cite{huang2022bevdet4d} fuses BEV features from different time sequences by coordinate alignment. BEVformer~\cite{li2022bevformer} is a transformer-based 3D object detection model that uses BEV to represent the scene and exploits multiple frames of images for spatial-temporal information fusion. After all, temporal fusion is an effective technique that enhances the continuity and relevance among different frames, and it can utilize the information from multiple frames to enrich the feature representation of each frame.

%% file: sec/3_Method.tex
\graphicspath{{\subfix{figures/}}}

\section{Method}
The overall architecture of GAFusion is illustrated in Fig.~\ref{Fig.main}. We feed the LiDAR point clouds and the corresponding multi-view images into the backbone to extract dual-stream features. The LiDAR stream first uses additional downsampling and sparse depth compression to obtain BEV features(Sec.~\ref{sec3.1}). The design of LiDAR guidance, which includes sparse depth guidance (SDG) and LiDAR occupancy guidance (LOG), is detailed in Sec.~\ref{sec3.2}. After LiDAR guidance, we adopt multi-scale dual-path transformer (MSDPT) module to enlarge the receptive fields of camera features (Sec.~\ref{sec3.3}). Then, the proposed LiDAR-guided adaptive fusion transformer (LGAFT) module is utilized to fuse different modalities of the BEV features (Sec.~\ref{sec3.4}). We also introduce the temporal fusion module to appropriately fuse the information from the previous frame (Sec.~\ref{sec3.5}).
\subsection{LiDAR and Camera Features Extraction} \label{sec3.1}
In the high-level feature extraction stage, we adopt a dual-stream approach to process the LiDAR point clouds and the multi-view images separately.

\textbf{For the LiDAR stream}, a common method~\cite{liangbevfusion, liu2022bevfusion, bai2022transfusion} is to use 3D sparse convolution~\cite{yan2018second} to extract single-scale features from the voxelized point clouds, which has a weak feature representation with limited receptive fields. Therefore, we use additional downsampling layers to compensate for this deficiency. The common sparse convolution features have strides of {1,2,4,8}, and the output sparse features are named {$F_{1}$, $F_{2}$, $F_{3}$, $F_{4}$} respectively. We adopt two additional downsampling layers with strides of {16,32} to obtain the {$F_{5}$, $F_{6}$} features. Finally, to effectively combine different scales of {$F_{4}$, $F_{5}$, $F_{6}$} features and maintain geometric and positional information, we use sparse depth compression to process different scales of the features. Specifically, we first align the spatial resolutions of {$F_{5}$, $F_{6}$} with $F_{4}$. For stage $i$, $F_{i}$ is a set of individual features. $p\in P_i$ is a position in 3D space, with the coordinate ($x_{p}$, $y_{p}$, $z_{p}$). In addition, we design a BEV grid of size ($x_p$,$y_p$) that only contains $P_c$, which aggregates the sparse features of different scales at the same height, and forms a rich BEV feature. The whole process is shown in Fig.~\ref{fig:AD,SHC}. Sparse features $F_{c}$ and their positions $P_{c}$ are obtained as follows:
\begin{equation}
\label{eq:combine}
\begin{aligned}
&{F}_{c}=F_4\cup(F_5\cup F_6),\qquad \\
&P_6' = \{(x_p \times 2^2, \, y_p \times 2^2, \, z_p \times 2^2) \,|\, p\in P_6\} \\
&P_5' = \{(x_p \times 2^1, \, y_p \times 2^1, \, z_p \times 2^1) \,|\, p\in P_5\} \\
&{P}_{c}=P_4\cup(P_5'\cup P_6').\qquad \\
\end{aligned}
\end{equation}

\textbf{For the camera stream}, Following the previous works~\cite{liangbevfusion, liu2022bevfusion}, we input multi-view images into backbone to obtain 2D image features $F_c\in\mathbb R^{N_c \times C \times H \times W}$ with sufficient semantic information, where $N_c$, $C$, $H$, $W$ denote the number of cameras, feature size, image height and image width respectively.
\subsection{LiDAR Guidance} \label{sec3.2}
\begin{figure}[t]
  \centering
    \includegraphics[width=\linewidth]{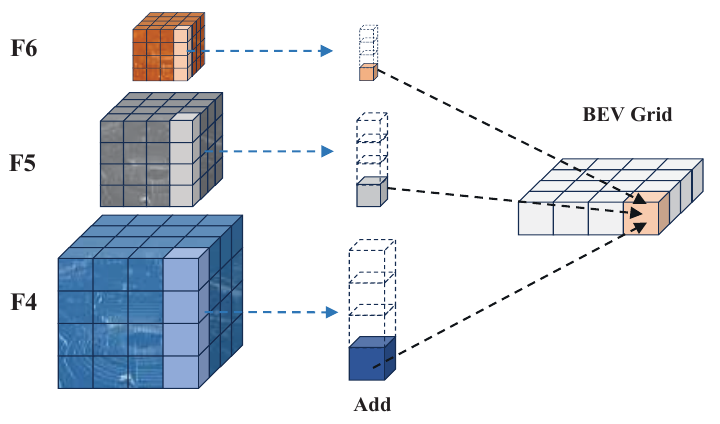}
   \caption{Additional downsampling and sparse height compression. This operation enlarges the receptive fields of the features and reduces the computational cost.}
   \label{fig:AD,SHC}
\end{figure}

 To integrate the camera high-level features into a unified BEV space, a view transformation is required, which first needs to project the 2D image features into 3D space. During this process, it is often difficult to estimate the depth distribution accurately, resulting in the loss of a lot of useful information in the BEV feature generated by the camera stream. To obtain a reliable depth distribution, our proposed LiDAR guidance consists of two parts: sparse depth guidance (SDG) and LiDAR occupancy guidance (LOG). They can help the image features better capture accurate geometric and depth information.
 
\begin{figure}[t]
  \centering
    \includegraphics[width=\linewidth]{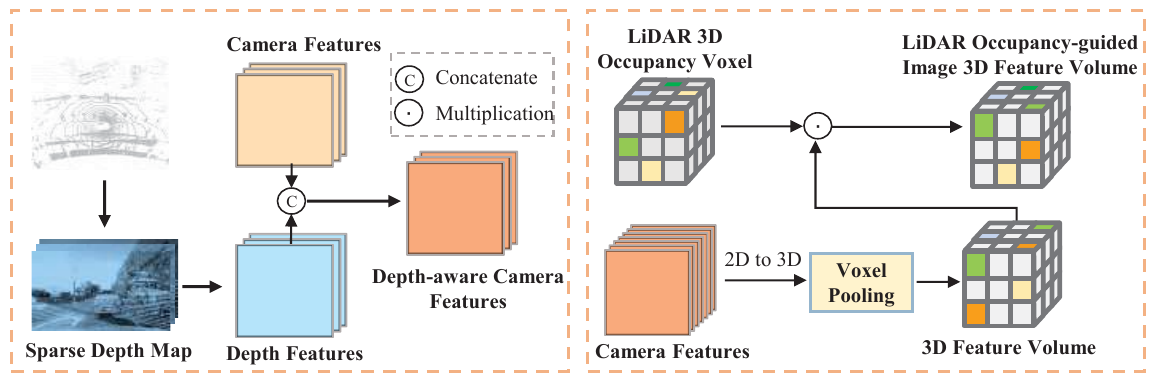}
   \caption{The architecture of sparse depth guidance (SDG) and LiDAR occupancy guidance (LOG). These two modules guide the 2D camera features to generate 3D features that contain sufficient semantic information and accurate depth information.}
   \label{fig:SOG+LOG}
\end{figure}

\textbf{Sparse Depth Guidance}. As shown in Fig.~\ref{fig:SOG+LOG}, SDG first projects each point of the input LiDAR point clouds into multi-view images, and obtains sparse multi-view depth maps. Then, they are fed into a shared encoder to extract depth features, which are concatenated with image features to form the depth-aware camera features. They are used as the input of view transformation, and finally voxel pooling~\cite{deng2021voxel} is employed to generate the image 3D feature volume, which is denoted as $F_c'$, $F_c'\in\mathbb R^{C \times Z \times H \times W}$. SDG can effectively incorporate LiDAR depth information and generate more accurate and reliable depth.

\textbf{LiDAR Occupancy Guidance}. Due to the sparsity and measurement noises of LiDAR point clouds, the depth information of some pixels is inaccurate. Therefore, LOG is proposed to address the aforementioned drawbacks, as shown in Fig.~\ref{fig:SOG+LOG}. Specifically, we first map LiDAR BEV features to 3D space to obtain the 3D features, and then attach an occupancy prediction head that estimates occupancy states to obtain the LiDAR 3D occupancy voxel, denoted as $O_L\in\mathbb R^{1 \times Z \times H \times W}$. It is worth noting that the resolution of $O_L$ is the same as that of $F_c'$. The LiDAR 3D occupancy voxel is then multiplied by $F_c'$ to obtain the LiDAR occupancy-guided image 3D feature volume $F_c''\in\mathbb R^{C \times Z \times H \times W}$ using the following equation:
\begin{equation}
\begin{aligned}
{F}_c''=Mul(F_c', \, O_L) 
\end{aligned}
\end{equation}

Where $Mul$ denotes element-wise multiplication with broadcasting operation. With the above designs, the 2D camera features contain sufficient semantic information and accurate depth information, which provide an excellent reference for subsequent module interactions.
\subsection{Multi-Scale Dual-Path Transformer} \label{sec3.3}
\begin{figure}[t]
  \centering
    \includegraphics[width=\linewidth]{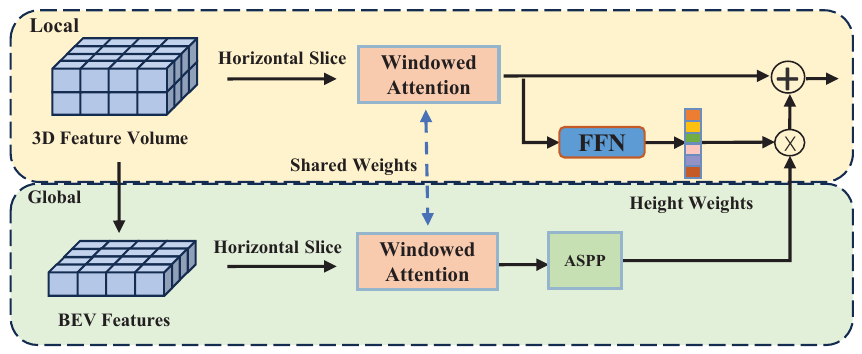}
   \caption{The schema of dual-path transformer (DPT), which effectively aggregates semantic information and expands the receptive fields of the camera features.}
   \label{fig:DPT}
\end{figure}

To effectively aggregate semantic information and enlarge the receptive fields, we improved a multi-scale dual-path transformer (MSDPT), a module inspired by OccFormer~\cite{zhang2023occformer}. Dual-path transformer (DPT) consists of a local path and a global path, which uses 3D convolution to perform downsampling to obtain features of different scales. The detailed structures of DPT are shown in Fig.~\ref{fig:DPT}. The local path is mainly used to extract fine-grained semantic structures. Since the height direction has less variation in 3D object detection, the local path only slices and processes the 3D feature volume extracted from the multi-view images in parallel along the horizontal direction. The global path attempts to acquire the semantic layout of the scene accurately. It first obtains BEV features by average pooling along the height dimension, and then interacts with the basic information of the BEV features. To improve computational efficiency, they both use windowed self-attention~\cite{liu2021swin}, and share weights. Finally, the 3D feature volume from the local path merges the sufficient semantic features from the global path. The dual-path outputs are $F_{local}\in\mathbb R^{C \times X \times Y \times Z}$ and $F_{global}\in\mathbb R^{C \times X \times Y}$, the combined output $F_{out}$ is computed as:
\begin{equation}
\begin{aligned}
{F}_\text{out} = {F}_\text{local} + \sigma({W_H} {F}_\text{local}) \cdot \text{unsqueeze} ({F}_\text{global}, -1)
\end{aligned}
\end{equation}

where $W_H$ refers to the aggregation weights along the height dimension generated by the FFN, $\sigma(\cdot)$ is the sigmoid function, and “unsqueeze” expands the global 2D features along the height.

\subsection{LiDAR-Guided Adaptive Fusion Transformer} \label{sec3.4}
\begin{figure}[t]
  \centering
    \includegraphics[width=\linewidth]{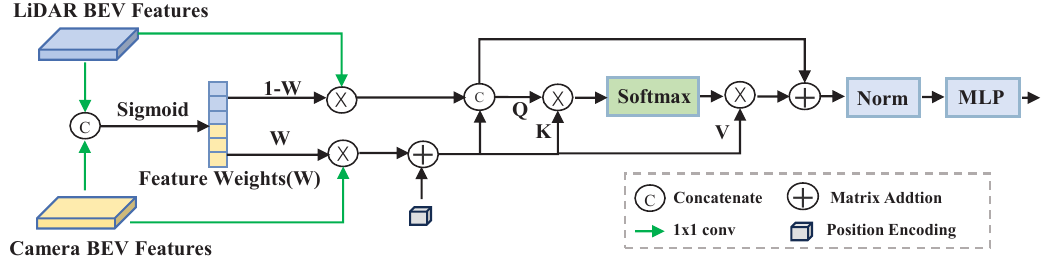}
   \caption{The overview of LiDAR-guided adaptive fusion transformer (LGAFT). LGAFT adaptively enhances the interaction between LiDAR and camera BEV features from a global perspective.}
   \label{fig:LGAFT}
\end{figure}

Recent works~\cite{liangbevfusion, liu2022bevfusion} simply concatenate different modalities of BEV features to obtain a shared BEV representation, which does not consider the information interaction and global spatial relevance among different modalities. To this end, LGAFT is developed to adaptively enhance the interaction of LiDAR BEV features $F_{LB}$ and camera BEV features $F_{CB}$ from a global perspective. The detailed architecture is illustrated in Fig.~\ref{fig:LGAFT}. We use $1\times1$ convolution to expand $F_{LB}$ and $F_{CB}$ to appropriate channels, and concatenate the expand BEV features $F_{LB}'$ and $F_{CB}'$ to obtain feature weights $W_F$ from a sigmoid function. Then, we adopt $W_F$ to fuse the LiDAR and camera BEV features adaptively, and the fused features are denoted as $F_{a}$. The weights $W_F$ can be expressed as:

\begin{equation}
\label{eq:combine}
\begin{aligned}
&F_{LB}'=\textrm{conv}_{1\times1}(F_{LB})\\
&F_{CB}'=\textrm{conv}_{1\times1}(F_{CB})\\
&W_F=\sigma(\textrm{Concat}(F_{LB}', F_{CB}'))
\end{aligned}
\end{equation}

Where “Concat” denotes the concatenate operation. To reduce the computation cost, we do not use the multi-head attention module in transformer structure. specifically, we adopt $F_{a}$ as the query of the cross-attention module. The adaptive camera features are regarded as the keys and values to avoid the gradient explosion convergence problem. Therefore, the final fused features $F_{BEV}$ can be presented as:

\begin{equation} \label{eq.query_generation}
\begin{split}
    &Q=\textrm{Concat}((1-W_F){F_{LB}'}, W_F{(F_{CB}'+P)}) \textbf{W}_{Q} \\
    &K=W_F{(F_{CB}'+P)}\textbf{W}_{K}\\
    &V={(F_{CB}'+P)}\textbf{W}_{V}\\
    &F_{BEV} = {MLP}\big({LN}({{Softmax}(\frac{QK^T}{\sqrt{C}})V})\big)\\
\end{split}
\end{equation}

Where $Q$, $K$, and $V$ denote the query, key, and value. $W_Q$, $W_K$ and $W_V$ are learnable parameters, $P$ stands for the learnable position embedding, $LN$ means layer normalization and $MLP$ is the multi-layer perception block.
\subsection{Temporal Fusion Module} \label{sec3.5}
Temporal information is crucial for the visual system to understand the surrounding environment. Temporal information can better help detect the motion states of the objects and occluded objects. We follow the fusion scheme of BEVDet4D~\cite{huang2022bevdet4d} and store the BEV features of historical frames in a memory buffer, and fuse the BEV features of the previous frame at each time. The detailed operation can be found in~\cite{huang2022bevdet4d}. Finally, we feed the fused BEV features into the BEV encoder and detection head to obtain the final detection results.

%% file: sec/4_Experiments.tex
\graphicspath{{\subfix{figures/}}}
\input{table/test_result}

\section{Experiments}
\subsection{Dataset and Metrics} \label{sec4.1}
Similar to previous works~\cite{liangbevfusion, liu2022bevfusion, bai2022transfusion}, we conduct extensive synthetic experiments on the nuScenes dataset. The nuScenes~\cite{caesar2020nuscenes} dataset is a large-scale autonomous driving benchmark, which includes 1000 scenes with images from 6 cameras with surrounding views, points from 5 Radars and 1 LiDAR. The scenes are officially split into 700/150/150 scenes for training/validation/testing. Each scene lasts for about 20 seconds, where key frames are annotated at 2Hz. Each frame of point cloud data corresponds to 6 RGB images with 360° horizontal FOV.

For the 3D detection task, we adopt the nuScenes Detection Score (NDS) and mean Average Precision (mAP) to evaluate the performance of the proposed model. In addition, the evaluation metrics of nuScenes also include five True Positive (TP) metrics, namely mean Average Translation Error (mATE), mean Average Scale Error (mASE), mean Average Orientation Error (mAOE), mean Average Velocity Error (mAVE) and mean Average Attribute Error (mAAE), which assess the performance of the model from different perspectives. NDS is the weighted average of mAP and five TP metrics.
\begin{figure*}[htbp]
\centering
\includegraphics[width=0.95\textwidth]{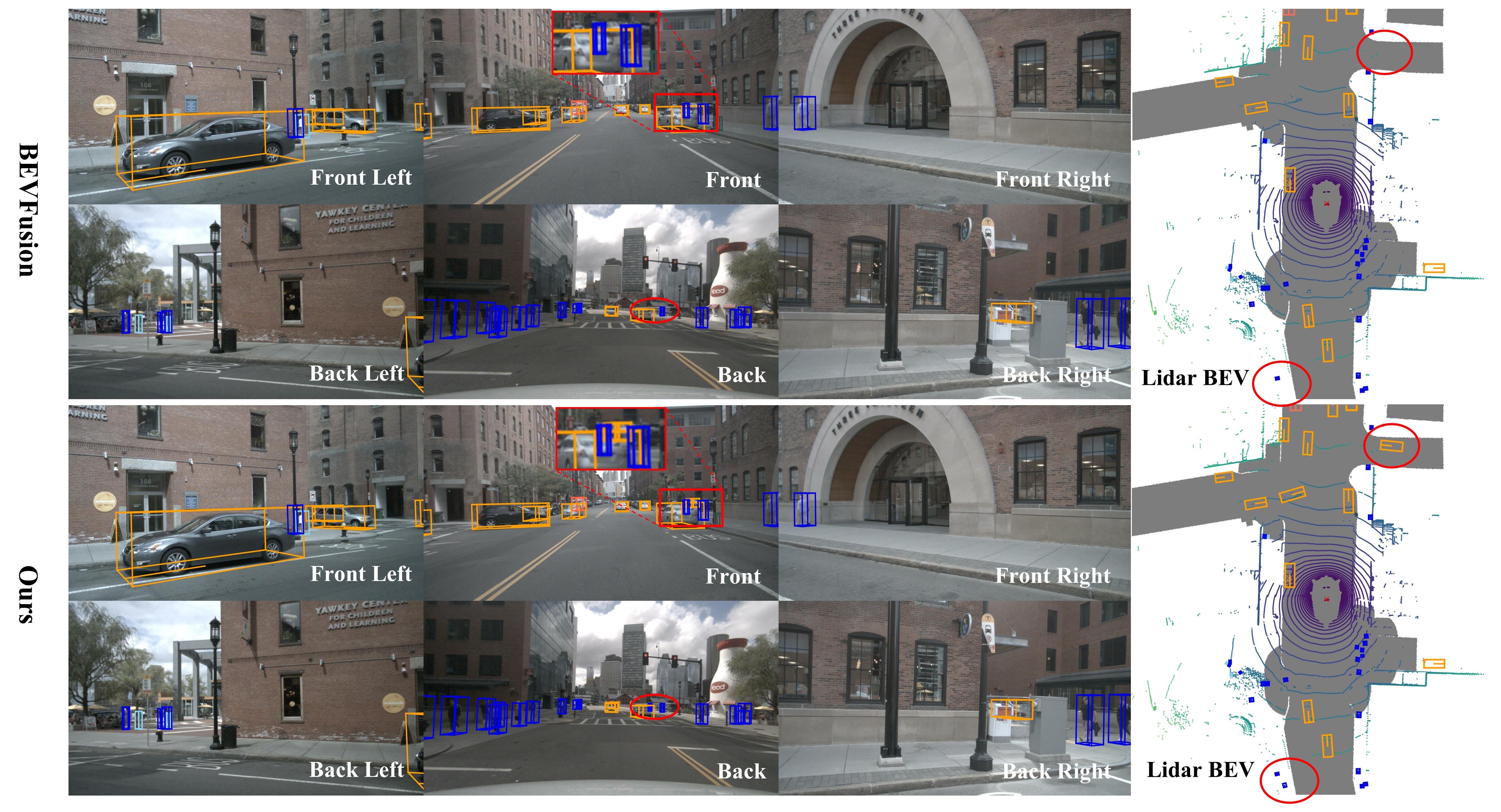}
\caption{Visualization results of BEVFusion and GAFusion on the nuScenes validation set. The red circles and boxes show the detection ability of GAFusion for small and occluded objects. }
\label{fig.visual} 
\end{figure*}
\input{table/val_result}
\subsection{Implementation Details} \label{sec4.2}
The developed model is implemented based on the MMDetection3D~\cite{mmdet3d2020} framework. For the LiDAR stream, we utilize the additional downsampling operation on top of VoxelNet~\cite{zhou2018voxelnet} as our backbone. For the camera stream, we adopt Swin-T~\cite{liu2021swin} and FPN as the image backbone, and use the pre-trained model of Swin-T. Similar to most models, the image resolution is 448${\times}$800, and the voxel size is (0.075m, 0.075m, 0.2m). The whole training process follows the previous work~\cite{liangbevfusion, liu2022bevfusion}. Firstly, a LiDAR detector is trained as 3D backbone for 20 epochs. Then, we freeze the pre-trained LiDAR components and jointly train for another 6 epochs according to the proposed framework. 
During the training stage, we use $AdamW$~\cite{loshchilov2017decoupled} optimizer with an initial learning rate of 5${\times}$10$^{-5}$ and a weight decay of 10$^{-2}$. GAFusion is trained on two 3090 GPUs with batch size of 4. In the inference stage, we do not use test-time augmentation (TTA) or multi-model ensemble.
\subsection{Results and Comparison} \label{sec4.3}
As shown in Table~\ref{tab:val} and Table~\ref{tab:test}, we report the results of GAFusion on the nuScenes validation and test sets, and compare them with other state-of-the-art models. The results show that, on the test set, GAFusion surpasses all the existing methods with 73.6$\%$ mAP and 74.9$\%$ NDS, such as MSMDFusion~\cite{jiao2022msmdfusion} and CMT~\cite{yan2023cmt}. It also achieves excellent performance on the validation set. In addition, we also provide the visualization results of GAFusion and BEVFusion to demonstrate the superiority of the proposed method, and they can be seen in Fig.~\ref{fig.visual}. This is attributed to better guidance mechanisms, larger receptive fields and a more suitable fusion method.
\subsection{Ablation Studies} \label{sec4.4}
To demonstrate the effectiveness and rationality of GAFusion, we conduct comprehensive ablation studies for each of the proposed components.

 \textbf{Additional downsampling and sparse height compression.} To prove the validity and generalization of this module, we separately insert the developed module into TransFusion~\cite{bai2022transfusion} and BEVFusion~\cite{liu2022bevfusion}, as shown in Table~\ref{tab:AD_SHC}. We do not use any augmentation strategies or multi-model ensemble during testing. The results illustrate that it can significantly improve the performance of different models. It enhances 1.0$\%$ mAP and 0.6$\%$ NDS in TransFusion, and 0.8$\%$ mAP and 0.5$\%$ NDS in BEVFusion, which indicates that it can aggregate multi-scale information.

 \input{table/AD_SHC_study}
 
\textbf{Impacts of LiDAR guidance.} To demonstrate that the contributions of LiDAR guidance indeed improve the model performance, we introduce SDG and LOG into BEVFusion~\cite{liu2022bevfusion}. Table~\ref{tab:SDG_LOG} presents the impacts of different combinations of the guidance modules in BEVFusion. We observe that the model performance brings about 1.4$\%$ mAP and 0.8$\%$ NDS gain with both SDG and LOG. When neither SDG nor LOG module is used, the model scores drop significantly. It can be attributed to the lack of guidance in the camera stream, and results in unreliable depth information. Moreover, it realizes 0.8$\%$ mAP and 0.4$\%$ NDS with SDG alone and 1.0$\%$ mAP and 0.6$\%$ NDS with LOG alone. The interaction effects of LOG are more remarkable, so we conjecture that directly interacting among 3D features can provide sufficient located information. SDG and LOG play their respective roles: the former integrates sparse depth information into 2D image features, and the latter guides depth information in 3D feature volume, which enables the camera stream to obtain rich geometric information.
 
\begin{figure}[t]
  \centering
    \includegraphics[width=\linewidth]{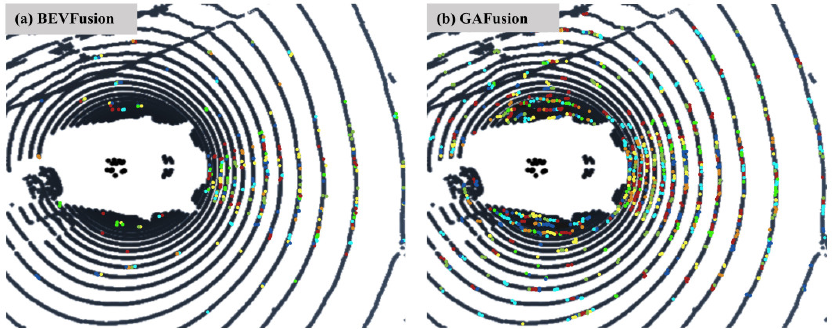}
   \caption{Receptive fields of the preliminary fused BEV features from different modalities. The colored dots indicate effective receptive fields.}
   \label{fig:efr}
\end{figure}

\input{table/SDG_LOG_study}

\textbf{BEV features fusion strategy.} We explore the impacts of different fusion methods, including addition, concatenation, LiDAR-guide fusion transformer (LGFT) and LGAFT. As shown in Table~\ref{tab:Fusion_strategy}, LGFT achieves a noticeable improvement over addition and concatenation, with about 0.7$\%$ mAP and 0.4$\%$ NDS. LGAFT further enhances 0.16$\%$ mAP and 0.11$\%$ NDS against LGFT due to the addition of adaptive mechanism. It presents that enhancing the interaction between LiDAR and camera BEV features from a global scope and the adaptive mechanism can sufficiently improve global spatial relevance.

\textbf{Effects of MSDPT.} We illustrate the related results to prove that MSDPT can effectively enlarge the receptive fields of camera features and aggregate semantic information. In Table~\ref{tab:MSDPT}, (1)-(5) are using output features (${F}_c''$) to fuse the features of different scales. Without MSDPT, the model performance drops by about 0.5$\%$ mAP and 0.4$\%$ NDS. Different scales of features also affect the model accuracy, which is due to the fact that the multi-scale operation can enlarge the receptive fields of camera features. However, redundant scales also cause too much computation and the performance enhancement is not obvious. Therefore, we select 3 scales to combine the different scale features for the balance of performance and computation.

\textbf{Larger receptive fields.} As shown in Fig.~\ref{fig:efr}, (a) and (b) illustrate the effective receptive fields of the fused features from the camera and LiDAR BEV features by BEVfusion~\cite{liu2022bevfusion} and GAFusion, respectively. We observe that GAFusion achieves larger effective receptive fields than BEVFusion. This is attributed to the additional downsampling and MSDPT modules, which indicate that multi-scale features can provide more contextual information. Besides, the global and local interaction of LGAFT contributes to enlarging the feature receptive fields to some extent.

\input{table/Fusionstrategy_Tem_study}
\input{table/MSDPTlayers_study}

\textbf{Temporal fusion.} In Table~\ref{tab:Fusion_strategy}, the temporal fusion improves about 0.3$\%$ mAP and 0.1$\%$ NDS. We integrate two frames of the BEV features, and the approach enables partial features alignment between adjacent frames, which leads to a marginal performance improvement. For multiple frames, it can attain higher enhancement, which is our future work.

%% file: table/test_result.tex
\begin{table*}[t]
\setlength{\tabcolsep}{2.6pt}
	\centering
    \label{tab:nuscenes_test}
	\begin{tabular}{l|c|cc|cccccccccc}
    \hline
         {Method} & {Modality} & {mAP$\uparrow$} & {NDS$\uparrow$} & {Car} & {Truck} & {C.V.} & {Trailer} & {Bus} & {Barrier} & {Motor.} & {Bike} & {Ped.} & {T.C.} \\
\hline
PointPillars~\cite{lang2019pointpillars} & L  & 30.5 & 45.3 & 68.4 & 23.0 & 4.1 & 23.4 & 28.2 & 38.9 & 27.4 & 1.1 & 59.7 & 30.8 \\
CenterPoint~\cite{yin2021center} & L  & 60.3 & 67.3 & 85.2 & 53.5 & 20.0 & 56.0 & 63.6 & 71.1 & 59.5 & 30.7 & 84.6 & 78.4 \\
TransFusion-L~\cite{bai2022transfusion} & L  & {65.5} & {70.2} & {86.2} & {56.7} & {28.2} & {58.8} & {66.3} & {78.2} & {68.3} & {44.2} & {86.1} & {82.0} \\
LargeKernel3D~\cite{chen2023largekernel3d} & L  & {65.3} & {70.5} & {85.9} & {55.3} & {26.8} & {60.2} & {66.2} & {74.3} & {72.5} & {46.6} & {85.6} & {80.0} \\
FocalFormer3D~\cite{chen2023focalformer3d} & L  & 68.7 & 72.6 & 87.8 & 59.4 & 37.8 & 65.7 & 73.0 & 77.8 & 77.4 & 52.4 & 90.0 & 83.4 \\
\hline
PointPainting~\cite{pointpainting} & LC   & 46.4 & 58.1 & 77.9 & 35.8 & 15.8 & 37.3 & 36.2 & 60.2 & 41.5 & 24.1 & 73.3 & 62.4 \\
3D-CVF~\cite{Yoo20203DCVFGJ} & LC   & 52.7 & 62.3 & 83.0 & 45.0 & 15.9 &  49.6& 48.8 & 65.9 & 51.2 & 30.4 & 74.2 & 62.9 \\
MVP~\cite{mvp} & LC  & 66.4 & 70.5  & 86.8 & 58.5 & 26.1 & 57.3 & 67.4 & 74.8 & 70.0 & 49.3 & {89.1} & 85.0 \\
TransFusion~\cite{bai2022transfusion} & LC  & {68.9} & {71.7} & 87.1 & 60.0 & {33.1} & 60.8 & 68.3 & {78.1} & 73.6 & 52.9 & {88.4} & 86.7 \\ 
AutoAlignV2~\cite{chen2022autoalignv2} & LC  & 68.4 & 72.4 & 87.0 & 59.0 & 33.1 & 59.3 & 69.3& - & 72.9 & 52.1 & 87.6 & - \\   
BEVFusion~\cite{liu2022bevfusion} &LC &70.2 &72.9  &88.6  &60.1  &39.3  & 63.8 &69.8  &80.0 &74.1 &51.0 &89.2 &85.2 \\        
BEVFusion~\cite{liangbevfusion} & LC  & {71.3} & {73.3} & 88.1 & 60.9 & 34.4 & 62.1 & 69.3 & 78.2 & 72.2 & 52.2 & 89.2 & 86.7 \\
CMT~\cite{yan2023cmt} & LC  & 70.4 & 73.0 & 87.2 & 61.5 & 37.5 & 62.8 & 72.4 &\textbf{ 86.9} & 79.4 & 58.3  &86.9 &83.1 \\          
DeepInteraction~\cite{yang2022deepinteraction} & LC  & 70.8 & 73.4 & 87.9 & 60.2 & 37.5 & 63.8 & 70.8 & {80.4} & 75.4 & 54.5 & 91.7 & 87.2 \\
FocalFormer3D~\cite{chen2023focalformer3d} & LC  & 71.6 & 73.9 & 88.5 & 61.4 & 35.9 & \textbf{66.4} & 71.7 & 79.3 & 80.3 & 57.1  &89.7 &85.3 \\
MSMDFusion~\cite{jiao2022msmdfusion} & LC  & 71.5 & 74.0 & 88.4 & 61.0 & 35.2 & {66.2} & 71.4 & {80.7} & 76.9 & 58.3  &90.6 &\textbf{88.1} \\
\hline
GAFusion(ours) & LC & {\textbf{73.6}} & {\textbf{74.9}} & \textbf{{89.4}} & \textbf{{65.3}} & \textbf{{42.4}} &{65.8}& \textbf{ {73.7} } & {79.2} & \textbf{{80.8}} & \textbf{{60.2}} & \textbf{{92.3}} & {87.0} \\
\hline
\end{tabular}
\captionsetup{width=.82\textwidth}
\caption{Comparison on the nuScenes test set. The models in the table are without ensemble or test-time augmentation. “L” is LiDAR, “C” is camera.}\vspace{-0.2cm}
\label{tab:test}
\end{table*}

%% file: table/val_result.tex
\begin{table}
  \centering
  \begin{adjustbox}{width=0.42\textwidth}
  \begin{tabular}{l|c|cc}
    \hline
         Methods & Modality & mAP$\uparrow$ & NDS$\uparrow$ \\
    \hline
    BEVFormer~\cite{li2022bevformer} & C & 41.6 & 51.7 \\
    PETRv2~\cite{liu2022petrv2} & C & 45.6 & 35.0 \\
    \hline
    CenterPoint~\cite{yin2021center} & L & 60.3 & 67.3 \\
    TransFusion-L~\cite{bai2022transfusion} & L & 65.5 & 70.2 \\
    \hline
    TransFusion~\cite{bai2022transfusion} & LC & 67.5 & 71.3 \\
    BEVFusion~\cite{liu2022bevfusion} & LC & 68.5 & 71.4 \\
    CMT~\cite{yan2023cmt} & LC & 67.9 & 70.8 \\
    MSMDFusion~\cite{jiao2022msmdfusion} & LC & 69.3 & 72.1 \\
    DeepInteration~\cite{yang2022deepinteraction} & LC & 69.9 & 72.6 \\
    SparseFusion~\cite{xie2023sparsefusion} & LC & 70.5 & 72.8 \\
    
    \hline
    GAFusion(ours) & LC & \textbf{72.1} & \textbf{73.5} \\  
    \hline
  \end{tabular}
  \end{adjustbox}
  \caption{Comparison on the nuScenes val set. The models in the table are without ensemble or test-time augmentation. }\vspace{-0.2cm}
  \label{tab:val}
\end{table}

%% file: table/AD_SHC_study.tex
\begin{table}
  \centering
  \begin{adjustbox}{width=0.45\textwidth}
  \begin{tabular}{@{}lllllccccc@{}}
    \toprule
    \vcenterhead{Backbone + Sparse2Dense} & \multicolumn{2}{c}{TransFusion} & \multicolumn{2}{c}{BEVFusion} \\
 \cmidrule(l{.5\tabcolsep}r{.5\tabcolsep}){2-3}
 \cmidrule(l{.5\tabcolsep}r{.5\tabcolsep}){4-5}
 & mAP$\uparrow$ & NDS$\uparrow$ & mAP$\uparrow$ & NDS$\uparrow$ \\
\midrule
Voxelnet + HC & 68.9  & 71.6 & 70.2 & 72.9 \\
\midrule
VoxelNet, AD + SHC & 69.9 & 72.4 & 71.0 & 73.4 \\
    \bottomrule
  \end{tabular}
  \end{adjustbox}
  \caption{Performance and generalization of additional downsampling (AD) and sparse height compression (SHC) on other common models. NDS/mAP comparison on nuScenes test set. “HC” is height compression.}\vspace{-0.2cm}
  \label{tab:AD_SHC}
\end{table}

%% file: table/SDG_LOG_study.tex
\begin{table}
  \centering
  \begin{adjustbox}{width=0.3\textwidth}
  \begin{tabular}{@{}lccccc@{}}
    \toprule
    & {SDG} & {LOG} & {mAP$\uparrow$} & {NDS$\uparrow$}  \\
\midrule
(1) & & & 68.52 & 71.38\\
(2) & \checkmark &  &  69.33 & 71.76\\
(3) &  &  \checkmark & 69.49 & 72.04 \\
(4) & \checkmark & \checkmark  & \textbf{69.93} & \textbf{72.24}\\
    \bottomrule
  \end{tabular}
  \end{adjustbox}
  \caption{Ablation study of LiDAR guidance on nuScenes val set. (1) represents the performance of the original BEVFusion model.}\vspace{-0.2cm}
  \label{tab:SDG_LOG}
\end{table}

%% file: table/Fusionstrategy_Tem_study.tex
\begin{table}
  \centering
  \begin{adjustbox}{width=0.32\textwidth}
  \begin{tabular}{@{}lccccc@{}}
    \toprule
     
& \multicolumn{2}{c}{BEV Fusion} & {Tem} & {mAP$\uparrow$} & {NDS$\uparrow$}  \\
\midrule
 \vcenterhead{(1)} & \multicolumn{2}{c}{\vcenterhead{ADD.}} &  & 70.85 & 72.82\\
 & \multicolumn{2}{l}{}  &  \checkmark  & 71.17 & 72.93   \\
\hdashline

\vcenterhead{(2)}  & \multicolumn{2}{c}{\vcenterhead{Cat.}} &  & 70.92 & 72.88\\
 & \multicolumn{2}{l}{}  &  \checkmark  & 71.21 & 72.98  \\
 \hdashline

\vcenterhead{(3)} & \multicolumn{2}{c}{\vcenterhead{LGFT}} &  &  71.63 & 73.32\\
 & \multicolumn{2}{l}{}  &  \checkmark  & 71.91 & 73.41  \\
 \hdashline

\vcenterhead{(4)}  & \multicolumn{2}{c}{\vcenterhead{LGAFT}} &  & 71.79 & 73.43\\
 & \multicolumn{2}{l}{}  &  \checkmark  & \textbf{72.08} & \textbf{73.53}  \\

    \bottomrule
  \end{tabular}
  \end{adjustbox}
  \caption{Ablation study of BEV fusion strategy and temporal fusion on nuScenes val set. “Tem” is temporal fusion. }\vspace{-0.2cm}
  \label{tab:Fusion_strategy}
\end{table}

%% file: table/MSDPTlayers_study.tex
\begin{table}
  \centering
  \begin{adjustbox}{width=0.35\textwidth}
  \begin{tabular}{@{}lccccccc@{}}
    \toprule
    & {C1} & {C2}& {C3} & {C4} & {mAP$\uparrow$} & {NDS$\uparrow$}  \\
\midrule
(1) &  & & & &  71.60 & 73.11\\
(2) & \checkmark &  &  & &   71.92 & 73.39\\
(3) &  & \checkmark &  & &  72.01 & 73.45 \\
(4) &  &  & \checkmark & &  \textbf{72.08} & 73.53\\
(5) &  &  &  & \checkmark &  72.07 & \textbf{73.54}\\
    \bottomrule
  \end{tabular}
  \end{adjustbox}
  \caption{Ablation study of MSDPT on nuScenes val set. C1-C4 denote the number of 3D convolution layers (1-4) applied to the 3D feature volume, respectively.}\vspace{-0.2cm}
  \label{tab:MSDPT}
\end{table}

%% file: sec/5-Conclusion.tex
\section{Conclusion}
We propose GAFusion, a more effective 3D object detection method in BEV representation, which is equipped with excellent guidance and fusion mechanisms. Additional downsampling and MSDPT are developed to enlarge the receptive fields of different modal features. Then SDG and LOG are employed to transform the 2D camera features into 3D features with sufficient located and spatial information. Afterward, we propose LGAFT to facilitate the fusion of LiDAR and camera BEV features. Finally, a temporal fusion module is adopted to aggregate features from different frames. Extensive experiments demonstrate the effectiveness and generality of our developed modules and GAFusion achieves state-of-the-art performances on the nuScenes dataset. We hope that the proposed components of GAFusion could provide more insights for subsequent research in this field.

\noindent\textbf{Acknowledgments} This project is developed based on the following open-sourced projects:MMDetection3D~\cite{mmdet3d2020}, TransFusion~\cite{bai2022transfusion}, BEVFusion~\cite{liu2022bevfusion}, BEVFusion~\cite{liangbevfusion}, OccFormer~\cite{zhang2023occformer}. Thanks for their excellent work.

%% file: main.bbl
\begin{thebibliography}{48}
\providecommand{\natexlab}[1]{#1}
\providecommand{\url}[1]{\texttt{#1}}
\expandafter\ifx\csname urlstyle\endcsname\relax
  \providecommand{\doi}[1]{doi: #1}\else
  \providecommand{\doi}{doi: \begingroup \urlstyle{rm}\Url}\fi

\bibitem[Bai et~al.(2022)Bai, Hu, Zhu, Huang, Chen, Fu, and Tai]{bai2022transfusion}
Xuyang Bai, Zeyu Hu, Xinge Zhu, Qingqiu Huang, Yilun Chen, Hongbo Fu, and Chiew-Lan Tai.
\newblock Transfusion: Robust lidar-camera fusion for 3d object detection with transformers.
\newblock In \emph{Proceedings of the IEEE/CVF Conference on Computer Vision and Pattern Recognition}, pages 1090--1099, 2022.

\bibitem[Caesar et~al.(2020)Caesar, Bankiti, Lang, Vora, Liong, Xu, Krishnan, Pan, Baldan, and Beijbom]{caesar2020nuscenes}
Holger Caesar, Varun Bankiti, Alex~H Lang, Sourabh Vora, Venice~Erin Liong, Qiang Xu, Anush Krishnan, Yu Pan, Giancarlo Baldan, and Oscar Beijbom.
\newblock nuscenes: A multimodal dataset for autonomous driving.
\newblock In \emph{Proceedings of the IEEE/CVF conference on computer vision and pattern recognition}, pages 11621--11631, 2020.

\bibitem[Cai et~al.(2023)Cai, Zhang, Zhou, Li, Ding, and Zhao]{cai2023bevfusion4d}
Hongxiang Cai, Zeyuan Zhang, Zhenyu Zhou, Ziyin Li, Wenbo Ding, and Jiuhua Zhao.
\newblock Bevfusion4d: Learning lidar-camera fusion under bird's-eye-view via cross-modality guidance and temporal aggregation.
\newblock \emph{arXiv preprint arXiv:2303.17099}, 2023.

\bibitem[Chen et~al.(2017)Chen, Ma, Wan, Li, and Xia]{chen2017mv3d}
Xiaozhi Chen, Huimin Ma, Ji Wan, Bo Li, and Tian Xia.
\newblock Multi-view 3d object detection network for autonomous driving.
\newblock In \emph{Proceedings of the IEEE conference on Computer Vision and Pattern Recognition}, pages 1907--1915, 2017.

\bibitem[Chen et~al.(2023{\natexlab{a}})Chen, Liu, Zhang, Qi, and Jia]{chen2023largekernel3d}
Yukang Chen, Jianhui Liu, Xiangyu Zhang, Xiaojuan Qi, and Jiaya Jia.
\newblock Largekernel3d: Scaling up kernels in 3d sparse cnns.
\newblock In \emph{Proceedings of the IEEE/CVF Conference on Computer Vision and Pattern Recognition}, pages 13488--13498, 2023{\natexlab{a}}.

\bibitem[Chen et~al.(2023{\natexlab{b}})Chen, Yu, Chen, Lan, Anandkumar, Jia, and Alvarez]{chen2023focalformer3d}
Yilun Chen, Zhiding Yu, Yukang Chen, Shiyi Lan, Anima Anandkumar, Jiaya Jia, and Jose~M Alvarez.
\newblock Focalformer3d: Focusing on hard instance for 3d object detection.
\newblock In \emph{Proceedings of the IEEE/CVF International Conference on Computer Vision}, pages 8394--8405, 2023{\natexlab{b}}.

\bibitem[Chen et~al.(2022)Chen, Li, Zhang, Fang, Jiang, and Zhao]{chen2022autoalignv2}
Zehui Chen, Zhenyu Li, Shiquan Zhang, Liangji Fang, Qinhong Jiang, and Feng Zhao.
\newblock Autoalignv2: Deformable feature aggregation for dynamic multi-modal 3d object detection.
\newblock \emph{arXiv preprint arXiv:2207.10316}, 2022.

\bibitem[Contributors(2020)]{mmdet3d2020}
MMDetection3D Contributors.
\newblock {MMDetection3D: OpenMMLab} next-generation platform for general {3D} object detection.
\newblock \url{https://github.com/open-mmlab/mmdetection3d}, 2020.

\bibitem[Deng et~al.(2021)Deng, Shi, Li, Zhou, Zhang, and Li]{deng2021voxel}
Jiajun Deng, Shaoshuai Shi, Peiwei Li, Wengang Zhou, Yanyong Zhang, and Houqiang Li.
\newblock Voxel r-cnn: Towards high performance voxel-based 3d object detection.
\newblock In \emph{Proceedings of the AAAI Conference on Artificial Intelligence}, pages 1201--1209, 2021.

\bibitem[Huang and Huang(2022)]{huang2022bevdet4d}
Junjie Huang and Guan Huang.
\newblock Bevdet4d: Exploit temporal cues in multi-camera 3d object detection.
\newblock \emph{arXiv preprint arXiv:2203.17054}, 2022.

\bibitem[Huang et~al.(2021)Huang, Huang, Zhu, Ye, and Du]{huang2021bevdet}
Junjie Huang, Guan Huang, Zheng Zhu, Yun Ye, and Dalong Du.
\newblock Bevdet: High-performance multi-camera 3d object detection in bird-eye-view.
\newblock \emph{arXiv preprint arXiv:2112.11790}, 2021.

\bibitem[Huang et~al.(2023)Huang, Zheng, Zhang, Zhou, and Lu]{huang2023tri}
Yuanhui Huang, Wenzhao Zheng, Yunpeng Zhang, Jie Zhou, and Jiwen Lu.
\newblock Tri-perspective view for vision-based 3d semantic occupancy prediction.
\newblock In \emph{Proceedings of the IEEE/CVF Conference on Computer Vision and Pattern Recognition}, pages 9223--9232, 2023.

\bibitem[Jiao et~al.(2023)Jiao, Jie, Chen, Chen, Ma, and Jiang]{jiao2022msmdfusion}
Yang Jiao, Zequn Jie, Shaoxiang Chen, Jingjing Chen, Lin Ma, and Yu-Gang Jiang.
\newblock Msmdfusion: Fusing lidar and camera at multiple scales with multi-depth seeds for 3d object detection.
\newblock In \emph{Proceedings of the IEEE/CVF Conference on Computer Vision and Pattern Recognition}, pages 21643--21652, 2023.

\bibitem[Ku et~al.(2018)Ku, Mozifian, Lee, Harakeh, and Waslander]{ku2018avod}
Jason Ku, Melissa Mozifian, Jungwook Lee, Ali Harakeh, and Steven~L Waslander.
\newblock Joint 3d proposal generation and object detection from view aggregation.
\newblock In \emph{2018 IEEE/RSJ International Conference on Intelligent Robots and Systems (IROS)}, pages 1--8. IEEE, 2018.

\bibitem[Lang et~al.(2019)Lang, Vora, Caesar, Zhou, Yang, and Beijbom]{lang2019pointpillars}
Alex~H Lang, Sourabh Vora, Holger Caesar, Lubing Zhou, Jiong Yang, and Oscar Beijbom.
\newblock Pointpillars: Fast encoders for object detection from point clouds.
\newblock In \emph{Proceedings of the IEEE/CVF conference on computer vision and pattern recognition}, pages 12697--12705, 2019.

\bibitem[Li et~al.(2023{\natexlab{a}})Li, Ge, Yu, Yang, Wang, Shi, Sun, and Li]{li2022bevdepth}
Yinhao Li, Zheng Ge, Guanyi Yu, Jinrong Yang, Zengran Wang, Yukang Shi, Jianjian Sun, and Zeming Li.
\newblock Bevdepth: Acquisition of reliable depth for multi-view 3d object detection.
\newblock In \emph{Proceedings of the AAAI Conference on Artificial Intelligence}, pages 1477--1485, 2023{\natexlab{a}}.

\bibitem[Li et~al.(2023{\natexlab{b}})Li, Yu, Choy, Xiao, Alvarez, Fidler, Feng, and Anandkumar]{li2023voxformer}
Yiming Li, Zhiding Yu, Christopher Choy, Chaowei Xiao, Jose~M Alvarez, Sanja Fidler, Chen Feng, and Anima Anandkumar.
\newblock Voxformer: Sparse voxel transformer for camera-based 3d semantic scene completion.
\newblock In \emph{Proceedings of the IEEE/CVF Conference on Computer Vision and Pattern Recognition}, pages 9087--9098, 2023{\natexlab{b}}.

\bibitem[Li et~al.(2021)Li, Wang, and Wang]{li2021lidarrcnn}
Zhichao Li, Feng Wang, and Naiyan Wang.
\newblock Lidar r-cnn: An efficient and universal 3d object detector.
\newblock In \emph{Proceedings of the IEEE/CVF Conference on Computer Vision and Pattern Recognition}, pages 7546--7555, 2021.

\bibitem[Li et~al.(2022)Li, Wang, Li, Xie, Sima, Lu, Qiao, and Dai]{li2022bevformer}
Zhiqi Li, Wenhai Wang, Hongyang Li, Enze Xie, Chonghao Sima, Tong Lu, Yu Qiao, and Jifeng Dai.
\newblock Bevformer: Learning bird’s-eye-view representation from multi-camera images via spatiotemporal transformers.
\newblock In \emph{Computer Vision--ECCV 2022: 17th European Conference, Tel Aviv, Israel, October 23--27, 2022, Proceedings, Part IX}, pages 1--18. Springer, 2022.

\bibitem[Li et~al.(2023{\natexlab{c}})Li, Yu, Austin, Fang, Lan, Kautz, and Alvarez]{li2023fbocc}
Zhiqi Li, Zhiding Yu, David Austin, Mingsheng Fang, Shiyi Lan, Jan Kautz, and Jose~M Alvarez.
\newblock Fb-occ: 3d occupancy prediction based on forward-backward view transformation.
\newblock \emph{arXiv preprint arXiv:2307.01492}, 2023{\natexlab{c}}.

\bibitem[Li et~al.(2023{\natexlab{d}})Li, Yu, Wang, Anandkumar, Lu, and Alvarez]{li2023fbbev}
Zhiqi Li, Zhiding Yu, Wenhai Wang, Anima Anandkumar, Tong Lu, and Jose~M Alvarez.
\newblock Fb-bev: Bev representation from forward-backward view transformations.
\newblock In \emph{Proceedings of the IEEE/CVF International Conference on Computer Vision}, pages 6919--6928, 2023{\natexlab{d}}.

\bibitem[Liang et~al.(2022)Liang, Xie, Yu, Xia, Lin, Wang, Tang, Wang, and Tang]{liangbevfusion}
Tingting Liang, Hongwei Xie, Kaicheng Yu, Zhongyu Xia, Zhiwei Lin, Yongtao Wang, Tao Tang, Bing Wang, and Zhi Tang.
\newblock Bevfusion: A simple and robust lidar-camera fusion framework.
\newblock \emph{Advances in Neural Information Processing Systems}, 35:\penalty0 10421--10434, 2022.

\bibitem[Liu et~al.(2022)Liu, Wang, Zhang, and Sun]{liu2022petr}
Yingfei Liu, Tiancai Wang, Xiangyu Zhang, and Jian Sun.
\newblock Petr: Position embedding transformation for multi-view 3d object detection.
\newblock In \emph{Computer Vision--ECCV 2022: 17th European Conference, Tel Aviv, Israel, October 23--27, 2022, Proceedings, Part XXVII}, pages 531--548. Springer, 2022.

\bibitem[Liu et~al.(2023{\natexlab{a}})Liu, Yan, Jia, Li, Gao, Wang, and Zhang]{liu2022petrv2}
Yingfei Liu, Junjie Yan, Fan Jia, Shuailin Li, Aqi Gao, Tiancai Wang, and Xiangyu Zhang.
\newblock Petrv2: A unified framework for 3d perception from multi-camera images.
\newblock In \emph{Proceedings of the IEEE/CVF International Conference on Computer Vision}, pages 3262--3272, 2023{\natexlab{a}}.

\bibitem[Liu et~al.(2021)Liu, Lin, Cao, Hu, Wei, Zhang, Lin, and Guo]{liu2021swin}
Ze Liu, Yutong Lin, Yue Cao, Han Hu, Yixuan Wei, Zheng Zhang, Stephen Lin, and Baining Guo.
\newblock Swin transformer: Hierarchical vision transformer using shifted windows.
\newblock In \emph{Proceedings of the IEEE/CVF international conference on computer vision}, pages 10012--10022, 2021.

\bibitem[Liu et~al.(2023{\natexlab{b}})Liu, Tang, Amini, Yang, Mao, Rus, and Han]{liu2022bevfusion}
Zhijian Liu, Haotian Tang, Alexander Amini, Xinyu Yang, Huizi Mao, Daniela~L Rus, and Song Han.
\newblock Bevfusion: Multi-task multi-sensor fusion with unified bird's-eye view representation.
\newblock In \emph{2023 IEEE international conference on robotics and automation (ICRA)}, pages 2774--2781. IEEE, 2023{\natexlab{b}}.

\bibitem[Loshchilov and Hutter(2017)]{loshchilov2017decoupled}
Ilya Loshchilov and Frank Hutter.
\newblock Decoupled weight decay regularization.
\newblock \emph{arXiv preprint arXiv:1711.05101}, 2017.

\bibitem[Ma et~al.(2022)Ma, Wang, Bai, Yang, Hou, Wang, Qiao, Yang, Manocha, and Zhu]{ma2022visionbev}
Yuexin Ma, Tai Wang, Xuyang Bai, Huitong Yang, Yuenan Hou, Yaming Wang, Yu Qiao, Ruigang Yang, Dinesh Manocha, and Xinge Zhu.
\newblock Vision-centric bev perception: A survey.
\newblock \emph{arXiv preprint arXiv:2208.02797}, 2022.

\bibitem[Miao et~al.(2023)Miao, Liu, Chen, Gong, Xu, Hu, and Zhou]{miao2023occdepth}
Ruihang Miao, Weizhou Liu, Mingrui Chen, Zheng Gong, Weixin Xu, Chen Hu, and Shuchang Zhou.
\newblock Occdepth: A depth-aware method for 3d semantic scene completion.
\newblock \emph{arXiv preprint arXiv:2302.13540}, 2023.

\bibitem[Philion and Fidler(2020)]{philion2020lift}
Jonah Philion and Sanja Fidler.
\newblock Lift, splat, shoot: Encoding images from arbitrary camera rigs by implicitly unprojecting to 3d.
\newblock In \emph{Computer Vision--ECCV 2020: 16th European Conference, Glasgow, UK, August 23--28, 2020, Proceedings, Part XIV 16}, pages 194--210. Springer, 2020.

\bibitem[Qi et~al.(2017)Qi, Su, Mo, and Guibas]{qi2017pointnet}
Charles~R Qi, Hao Su, Kaichun Mo, and Leonidas~J Guibas.
\newblock Pointnet: Deep learning on point sets for 3d classification and segmentation.
\newblock In \emph{Proceedings of the IEEE conference on computer vision and pattern recognition}, pages 652--660, 2017.

\bibitem[Schuster and Paliwal(1997)]{schuster1997birnn}
Mike Schuster and Kuldip~K Paliwal.
\newblock Bidirectional recurrent neural networks.
\newblock \emph{IEEE transactions on Signal Processing}, 45\penalty0 (11):\penalty0 2673--2681, 1997.

\bibitem[Shi et~al.(2020)Shi, Guo, Jiang, Wang, Shi, Wang, and Li]{pvrcnn}
Shaoshuai Shi, Chaoxu Guo, Li Jiang, Zhe Wang, Jianping Shi, Xiaogang Wang, and Hongsheng Li.
\newblock Pv-rcnn: Point-voxel feature set abstraction for 3d object detection.
\newblock In \emph{CVPR}, pages 10529--10538, 2020.

\bibitem[Shi et~al.(2022)Shi, Jiang, Deng, Wang, Guo, Shi, Wang, and Li]{pvrcnn++}
Shaoshuai Shi, Li Jiang, Jiajun Deng, Zhe Wang, Chaoxu Guo, Jianping Shi, Xiaogang Wang, and Hongsheng Li.
\newblock Pv-rcnn++: Point-voxel feature set abstraction with local vector representation for 3d object detection.
\newblock \emph{IJCV}, pages 1--21, 2022.

\bibitem[Sindagi et~al.(2019)Sindagi, Zhou, and Tuzel]{sindagi2019mvx}
Vishwanath~A Sindagi, Yin Zhou, and Oncel Tuzel.
\newblock Mvx-net: Multimodal voxelnet for 3d object detection.
\newblock In \emph{2019 International Conference on Robotics and Automation (ICRA)}, pages 7276--7282. IEEE, 2019.

\bibitem[Vora et~al.(2020)Vora, Lang, Helou, and Beijbom]{pointpainting}
Sourabh Vora, Alex~H Lang, Bassam Helou, and Oscar Beijbom.
\newblock Pointpainting: Sequential fusion for 3d object detection.
\newblock In \emph{Proceedings of the IEEE/CVF conference on computer vision and pattern recognition}, pages 4604--4612, 2020.

\bibitem[Wang et~al.(2023)Wang, Zhu, Xu, Zhang, Wei, Chi, Ye, Du, Lu, and Wang]{wang2023openoccupancy}
Xiaofeng Wang, Zheng Zhu, Wenbo Xu, Yunpeng Zhang, Yi Wei, Xu Chi, Yun Ye, Dalong Du, Jiwen Lu, and Xingang Wang.
\newblock Openoccupancy: A large scale benchmark for surrounding semantic occupancy perception.
\newblock In \emph{Proceedings of the IEEE/CVF International Conference on Computer Vision}, pages 17850--17859, 2023.

\bibitem[Wang et~al.(2022)Wang, Guizilini, Zhang, Wang, Zhao, and Solomon]{wang2022detr3d}
Yue Wang, Vitor~Campagnolo Guizilini, Tianyuan Zhang, Yilun Wang, Hang Zhao, and Justin Solomon.
\newblock Detr3d: 3d object detection from multi-view images via 3d-to-2d queries.
\newblock In \emph{Conference on Robot Learning}, pages 180--191. PMLR, 2022.

\bibitem[Xie et~al.(2023)Xie, Xu, Rakotosaona, Rim, Tombari, Keutzer, Tomizuka, and Zhan]{xie2023sparsefusion}
Yichen Xie, Chenfeng Xu, Marie-Julie Rakotosaona, Patrick Rim, Federico Tombari, Kurt Keutzer, Masayoshi Tomizuka, and Wei Zhan.
\newblock Sparsefusion: Fusing multi-modal sparse representations for multi-sensor 3d object detection.
\newblock In \emph{Proceedings of the IEEE/CVF International Conference on Computer Vision}, pages 17591--17602, 2023.

\bibitem[Yan et~al.(2023)Yan, Liu, Sun, Jia, Li, Wang, and Zhang]{yan2023cmt}
Junjie Yan, Yingfei Liu, Jianjian Sun, Fan Jia, Shuailin Li, Tiancai Wang, and Xiangyu Zhang.
\newblock Cross modal transformer: Towards fast and robust 3d object detection.
\newblock In \emph{Proceedings of the IEEE/CVF International Conference on Computer Vision}, pages 18268--18278, 2023.

\bibitem[Yan et~al.(2018)Yan, Mao, and Li]{yan2018second}
Yan Yan, Yuxing Mao, and Bo Li.
\newblock Second: Sparsely embedded convolutional detection.
\newblock \emph{Sensors}, 18\penalty0 (10):\penalty0 3337, 2018.

\bibitem[Yang et~al.(2022)Yang, Chen, Miao, Li, Zhu, and Zhang]{yang2022deepinteraction}
Zeyu Yang, Jiaqi Chen, Zhenwei Miao, Wei Li, Xiatian Zhu, and Li Zhang.
\newblock Deepinteraction: 3d object detection via modality interaction.
\newblock In \emph{NeurIPS}, 2022.

\bibitem[Yin et~al.(2021{\natexlab{a}})Yin, Zhou, and Kr{\"a}henb{\"u}hl]{mvp}
Tianwei Yin, Xingyi Zhou, and Philipp Kr{\"a}henb{\"u}hl.
\newblock Multimodal virtual point 3d detection.
\newblock \emph{Advances in Neural Information Processing Systems}, 34:\penalty0 16494--16507, 2021{\natexlab{a}}.

\bibitem[Yin et~al.(2021{\natexlab{b}})Yin, Zhou, and Krahenbuhl]{yin2021center}
Tianwei Yin, Xingyi Zhou, and Philipp Krahenbuhl.
\newblock Center-based 3d object detection and tracking.
\newblock In \emph{Proceedings of the IEEE/CVF conference on computer vision and pattern recognition}, pages 11784--11793, 2021{\natexlab{b}}.

\bibitem[Yoo et~al.(2020)Yoo, Kim, Kim, and Choi]{Yoo20203DCVFGJ}
Jin~Hyeok Yoo, Yeocheol Kim, Ji~Song Kim, and J. Choi.
\newblock {3D-CVF}: Generating joint camera and lidar features using cross-view spatial feature fusion for 3d object detection.
\newblock In \emph{ECCV}, 2020.

\bibitem[Zhai et~al.(2022)Zhai, Wang, Pan, Gao, and Hu]{zhai20223Dvid}
Zhenyu Zhai, Qiantong Wang, Zongxu Pan, Zhentong Gao, and Wenlong Hu.
\newblock Muti-frame point cloud feature fusion based on attention mechanisms for 3d object detection.
\newblock \emph{Sensors}, 22\penalty0 (19):\penalty0 7473, 2022.

\bibitem[Zhang et~al.(2023)Zhang, Zhu, and Du]{zhang2023occformer}
Yunpeng Zhang, Zheng Zhu, and Dalong Du.
\newblock Occformer: Dual-path transformer for vision-based 3d semantic occupancy prediction.
\newblock In \emph{Proceedings of the IEEE/CVF International Conference on Computer Vision}, pages 9433--9443, 2023.

\bibitem[Zhou and Tuzel(2018)]{zhou2018voxelnet}
Yin Zhou and Oncel Tuzel.
\newblock Voxelnet: End-to-end learning for point cloud based 3d object detection.
\newblock In \emph{Proceedings of the IEEE conference on computer vision and pattern recognition}, pages 4490--4499, 2018.

\end{thebibliography}
